\documentclass[conference]{IEEEtran}
\IEEEoverridecommandlockouts

\usepackage{amsmath,amssymb,amsfonts}
\usepackage{algorithmic}
\usepackage{graphicx,booktabs}
\usepackage{textcomp}
\usepackage[dvipsnames]{xcolor}
\usepackage{color}
\usepackage{colortbl}
\usepackage{multirow}
\usepackage{url}
\usepackage[pagebackref=false,breaklinks=true,colorlinks,bookmarks=false]{hyperref}
\usepackage{cleveref}

\usepackage[compress]{cite}

\renewcommand{\baselinestretch}{0.95}

\def\BibTeX{{\rm B\kern-.05em{\sc i\kern-.025em b}\kern-.08em
    T\kern-.1667em\lower.7ex\hbox{E}\kern-.125emX}}

\usepackage{fancyhdr}
\fancypagestyle{firststyle}
{
   \fancyhf{}
   \fancyhead[C]{
    \textcolor{red}{
       The content of this paper was published in IWBF 2023. If you find this research useful, please cite the following paper: \\ Ž. Babnik, N. Damer, V. Štruc, "Optimization-Based Improvement of Face Image Quality Assessment Techniques" in Proceedings of the IEEE International Workshop on Biometrics and Forensics (IWBF), 2023.}
    }
}

\begin{document}

\title{Optimization-Based Improvement of Face Image Quality Assessment Techniques\\
\thanks{Supported by ARRS: P2-0250(B), J2-2501(A), Junior Researcher grants.}\vspace{-2mm}
}

\author{\IEEEauthorblockN{Žiga Babnik\IEEEauthorrefmark{1}, Naser Damer\IEEEauthorrefmark{2}, Vitomir Štruc\IEEEauthorrefmark{1}}
\IEEEauthorblockA{\textit{\IEEEauthorrefmark{1}University of Ljubljana, Faculty of Electrical Engineering Tržaška cesta 25, 1000 Ljubljana, Slovenia}\\\textit{\IEEEauthorrefmark{2}Fraunhofer IGD, Darmstadt, Germany}} \vspace{-4mm}
{\small \url{https://github.com/LSIbabnikz/Optimization-Based-Improvement-of-FIQA-Techniques}}
}


\maketitle
\thispagestyle{firststyle}


\begin{abstract}
Contemporary face recognition~(FR) models achieve near-ideal recognition performance in constrained settings, yet do not fully translate the performance to unconstrained (real-world) scenarios. To help improve the performance and stability of FR systems in such unconstrained settings, face image quality assessment (FIQA) techniques try to infer sample-quality information from the input face images that can aid with the recognition process. 
While existing FIQA techniques are able to efficiently capture the differences between high and low quality images, they typically cannot fully distinguish between images of similar quality, leading to lower performance in many scenarios.
To address this issue, we present in this paper a supervised quality-label optimization approach, aimed at improving the performance of existing FIQA techniques. The developed optimization procedure infuses additional information (computed with a selected FR model) into the initial quality scores generated with a given FIQA technique to produce better estimates of the ``actual'' image quality. We evaluate the proposed approach in comprehensive experiments with six  state-of-the-art FIQA approaches ({CR-FIQA}, {FaceQAN}, {SER-FIQ}, {PCNet}, {MagFace}, {SDD-FIQA}) on five commonly used benchmarks ({LFW}, {CFP-FP}, {CPLFW}, {CALFW}, {XQLFW}) 
using three targeted FR models (ArcFace, ElasticFace, 
CurricularFace) with highly encouraging results.
\end{abstract}


\section{Introduction}

Modern face recognition~(FR) systems achieve excellent results even with large-scale recognition problems, as long as the appearance variability of the facial images is reasonably constrained. However, the performance in constrained scenarios does not always translate to real-world scenarios where out-of-distribution data, often of poor quality, still presents a challenge for the majority of existing FR models~\cite{fr1, fr2}. 

Face image quality assessment (FIQA) techniques aim to assist FR models in such challenging scenarios by providing additional information on the quality of facial images. This quality information can then be used to either reject low-quality samples that typically lead to false match errors or design robust quality-aware face recognition techniques. Thus, different from general purpose image quality assessment~(IQA) methods~\cite{iqa1, iqa2, iqa3} that commonly  measure the perceived visual quality of images by examining explicit image characteristics, such as sharpness, lighting conditions and resolution, FIQA techniques typically try to capture the utility (or fitness) of the given face image for the recognition task~\cite{isoiec}. 
In other words, 
they measure the usefulness of the sample for face recognition.

Several groups of FIQA techniques that differ slightly in their approach have been proposed so far in the literature \cite{surveypaper}. The majority of recent techniques learns quality-estimation networks using (reference) quality information inferred from a large database of face images~\cite{faceqnet, sddfiqa, lightqnet, pcnet}. Another notable group of FIQA techniques estimates quality based only on the information present in the input image and the characteristics of the targeted FR system~\cite{serfiq, faceqan}. More recently, approaches have also appeared that incorporate quality estimation directly into the FR process~\cite{pfe, magface}, paving the way towards quality-aware face recognition. 

\begin{figure*}[!ht]
    \centering
    \includegraphics[width=0.83\textwidth]{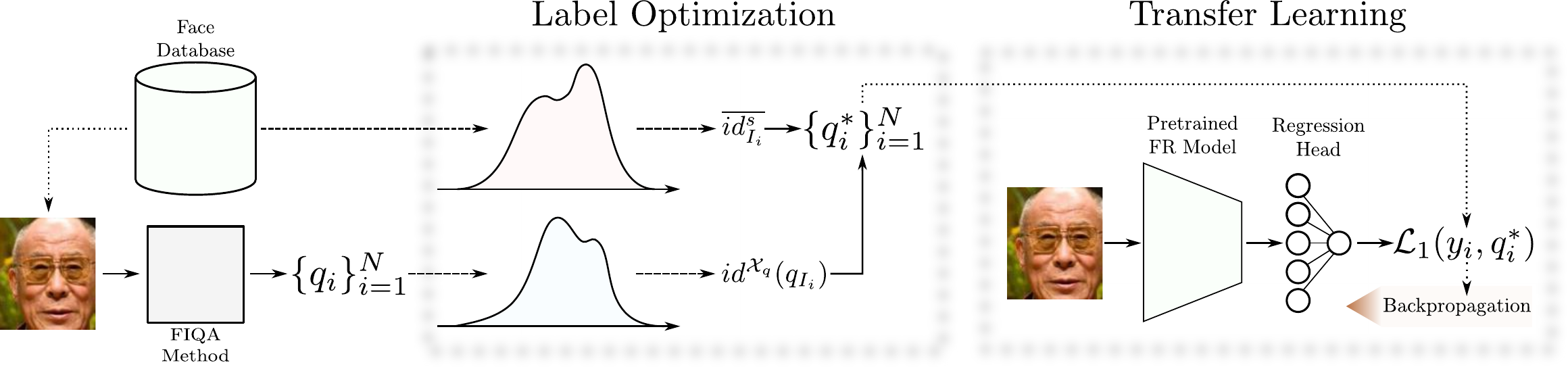}\vspace{-2mm}
    \caption{\textbf{Overview of the proposed method that consists of: Label Optimization and Transfer Learning.} The \textit{label-optimization} step incorporates information extracted from mated image pairs into quality scores precomputed with an existing FIQA technique. The \textit{transfer-learning} step is then used to train a FR model, extended with a regression head, on the optimized quality-scores. The learned regressor is finally utilized for quality estimation.}\vspace{-2mm}
    \label{fig:overview}
\end{figure*}

While most of the existing FIQA techniques perform well enough to distinguish between high-quality and low-quality facial images, correctly ranking face images of similar quality remains an open problem. The correct (optimal) ordering does not depend solely on the input face images, but also on the targeted FR model. 
Each model may, in a sense, \textit{perceive} the quality of individual samples differently due to different model-specific biases introduced by the learning process and the data used for training~\cite{bias1, bias2}.
This observation also suggests that FIQA techniques, that are not FR model specific, can not determine the correct order for all possible FR models. For this reason, we propose in this paper a novel optimization approach, that attempts to improve the predictive power of any given FIQA approach by incorporating quality information obtained by a particular FR model into the quality scores generated by the selected FIQA approach. Thus, the main contributions of this papers are:
\begin{itemize}

    \item{A novel optimization approach that incorporates model-specific quality information into the quality scores produced by existing FIQA techniques with the goal of improving FIQA performance. 
    }

    \item{An in-depth evaluation of the proposed optimization approach over six FIQA techniques, five datasets, three recognition models and in two settings that demonstrates significant performance gains in most situations. 
    }
\end{itemize}

\section{Related Work}

In this section, we briefly review previous FIQA research that can be broadly categorized into three groups: $(i)$ analytical, $(ii)$ regression and $(iii)$ model-based techniques. More in-depth information on face quality assessment can be found in the comprehensive survey paper by Schlett \textit{et al.}~\cite{surveypaper}.

\textbf{Analytical FIQA} techniques are mostly unsupervised and rely solely on the information that can be extracted directly from the given input sample. Techniques from this group typically focus on the visual quality of the facial images and, as a result, often exhibit limited performance. The method proposed by Gao \textit{et~al.}~\cite{anal1}, for example, attempts to extract quality information based on facial symmetry estimation only. Zhang \textit{et~al.}~\cite{illumination} try to quantify quality based on image illumination information, while Lijun \textit{et al.}~\cite{multi_anal} combine multiple cues, such as occlusions, blur and pose for the quality-estimation task. Different from these methods, two analytical FIQA techniques have been proposed recently that in addition to the characteristics of the input image also consider the targeted FR system during the quality estimation task. The first, SER-FIQ by Terhörst \textit{et~al.}~\cite{serfiq}, uses the properties of dropout layers to quantify quality, while FaceQAN, by Babnik \textit{et~al.}~\cite{faceqan}, exploits adversarial examples for quality assessment. Both methods were shown to yield state-of-the-art performance for various FR models and different benchmarks.

\textbf{Regression-based FIQA} techniques are the most numerous and usually learn a quality estimation (regression) model to predict quality scores based on some pseudo (ground-truth) quality labels. FaceQNet~\cite{faceqnet}, for example, trains a ResNet50 model using labels obtained by embedding comparisons with the highest quality image of each subject. Here, the highest quality images are determined using an external quality compliance tool. A similar approach, called PCNet~\cite{pcnet}, trains a quality-regression network on mated-image pairs, with the goal of predicting the similarity of the image pair. LightQNet~\cite{lightqnet} builds on the ideas introduced with PCNet, but additionally relies on a so-called Identification Quality~(IQ) loss, while SDD-FIQA~\cite{sddfiqa} considers both mated and non-mated similarity scores between a large number of samples to determine the final reference quality for the regression task.

\textbf{Model-based FIQA} techniques are less common and usually try to combine face recognition and quality assessment in a single quality-aware face recognition task. The main goal of these techniques is to simultaneously produce, for a given sample, its embedding and an estimate of the sample's quality. For example, the approach presented by Shi and Jain~\cite{pfe}, estimates a mean and variance vector for each given input sample, where the mean vector represents the embedding, while the variance provides the corresponding uncertainty and can be interpreted as a sample quality estimate. MagFace~\cite{magface}, a similar approach by Meng~\textit{et~al.}, uses a modified version of the commonly used ArcFace loss, called MagFace loss, which is able to generate quality-aware embeddings, by incorporating quality information into the magnitude of the embedding itself.

The method we propose cannot be clearly assigned to one of the above groups, because it relies on an already existing FIQA approach (from any of the three groups) to generate reference quality scores. In a sense, it distills FIQA knowledge from any existing technique. 
However, if treated as a black-box, the proposed FIQA approach can be thought of as a regression-based technique, as it trains a regression model using quality labels extracted from a large database.

\begin{figure*}
    \centering
    \includegraphics[width=0.81\textwidth]{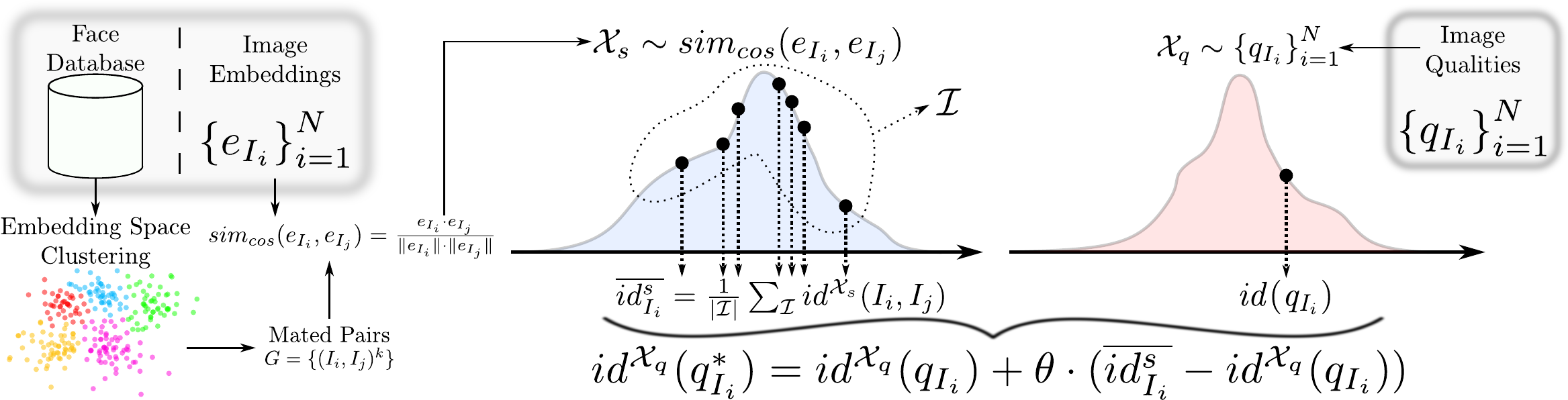}\vspace{-2mm}
    \caption{\textbf{Overview of Label Optimization.}
    We present a visualization of the proposed optimization scheme. Based on the embeddings $\{e_{I_i}\}^{N_k}_{i=1}$ we first generate mated image pairs. From the image pairs, we compute the pair similarity distribution $\mathcal{X}_s$ using the cosine similarity of the image embedings. At the same time, we also construct the quality distribution $\mathcal{X}_q$ from the given quality scores $\{q_{I_i}\}^N_{i=1}$. The mean similarity index $\overline{id^s_{I_i}}$, calculated as the average index of all image pairs from $\mathcal{I}$, is then used to update the quality index $id^{\mathcal{X}_q}(q_{I_i})$, using the equation presented above.}
    \label{fig:my_label}\vspace{-2mm}
\end{figure*}

\section{Methodology}

State-of-the-art FIQA techniques are able to efficiently discriminate between images of distinctly different qualities, yet may not be able to properly distinguish between images of similar quality. To exacerbate this problem, the relative ordering of images of similar quality may additionally depend on the targeted FR model, which not all FIQA techniques take into account. Because face quality assessment aims to quantify the utility of face images for a given FR model, the slight variations in the biases present in modern FR systems may result in different (optimal) quality scores for different FR models. For this reason, we propose in this paper an approach that aims to incorporate FR model-specific quality information into (some initial) quality scores, with the goal of improving the fine-grained performance of existing FIQA techniques. The overall pipeline of the proposed approach, shown in Fig.~\ref{fig:overview}, consists of two main steps: $(i)$ \textit{label optimization} and $(ii)$ \textit{transfer learning}. The label-optimization step aims to incorporate additional quality-related information into the baseline quality labels, precomputed using a selected (existing) FIQA approach. The optimized quality labels are then used in a transfer-learning scheme, that uses a pre-trained FR model, extended with a quality-regression head.

\subsection{Method Overview}

Let $Q$ and $M$ denote a given FIQA method and a pre-trained FR model that produce quality scores $q_I = Q(I)$ and embeddings $e_I = M(I)$, respectively, for an arbitrary input face image $I$, and $\{I_i\}^N_{i=1}$ denote a large facial image database consisting of $N$ distinct images. 
The goal of our approach is to train a regression-based quality-estimation model $Q^* = H(M(I))$, where $H$ represents a quality-regression head, that outperforms the initial FIQA method $Q$. The model $Q^*$ is trained on optimized quality labels $\{q^*_i\}^N_{i=1}$ generated by the proposed optimization scheme $O$. The method relies on information obtained from mated image pairs of the face database $\{I_i\}^N_{i=1}$. Details on the procedure are given below. 

\subsection{Initialization}

We first extract initial quality scores $q_i = Q(I_i)$ and embeddings $e_i = M(I_i)$ from all images of the given face image database $\{I_i\}^N_{i=1}$ using the selected FIQA method $Q$ and chosen FR model $M$. This initialization step is conducted once and provides the input data for the label optimization and consequently the transfer learning procedures.

\subsection{Label Optimization}

Looking at past research~\cite{faceqnet, sddfiqa, lightqnet, pcnet}, we observe that quality information is often inferred from mated image comparisons, where the term \textit{mated images} refers to two unique images of the same individual. We, therefore, follow this insight and use such information in our optimization approach as well. By computing the similarity of mated image pairs in the embedding space of the given FR model $M$, we are also able to include FR-specific quality estimates in the optimization. 

\textbf{Selecting mated image pairs.} Large-scale databases contain a significant amount of images for each individual, where many of the images may be nearly identical. Selecting all possible mated pairs, can, therefore, introduce database specific biases into our approach and adversely affect performance. 

To avoid such issues, we propose a technique for sampling mated image pairs based on clustering.
We use a clustering procedure to find groups of similar images and to identify the most informative (and least redundant) mated image pairs. We cluster the embedding space $\mathcal{E}^k = \{e_i^k\}^{N_k}_{i=1}$ corresponding to images of each individual $k=1,...,K$ present in the database using K-Means, where $N=\sum_k N_k$. The algorithm initializes $C$ cluster centers by randomly sampling the given data points and iteratively corrects them using nearby examples. 
For each image $I^k_c$ of the $k$-th individual belonging to cluster $c \in [1,C]$, we randomly select images from all other clusters $c' \neq c,\; c' \in [1,C]$ to form mated pairs $(I^k_c, I^k_{c'})$. By repeating this process for each image of every individual, we obtain the final mated image pairs for the label-optimization procedure $G = \{(I_i, I_j)^l\}_{l=1}^L$, where $i \neq j$ and $L = N \cdot (c - 1)$. 

\textbf{Optimizing prior quality scores.} 
We aim at optimizing the initial quality labels $\{q_{I_i}\}^N_{i=1}$ using information provided by the average pair similarity $sim_{I_i}$ of each image. In other words, if an image has a low quality score, yet its average pair similarity is high, we want to increase its quality. Conversely, if the opposite is true, we want to decrease it. The design of the optimization procedure is based on the assumption that the initial quality scores already provide a reasonable estimate of the true quality. 
We, therefore, try to retain the overall quality distribution over the face database. As a result, we simply rearrange the order of the images in the original quality score distribution generated by the selected FIQA technique $Q$ instead of computing new optimal quality scores that could differ significantly from the initial estimates.

From the list of genuine image pairs $G$, we first calculate the cosine similarity of all image embedding pairs, i.e.:
\begin{equation}
    sim_{cos}(e_{I_i}, e_{I_j}) = \frac{e_{I_i} \cdot e_{I_j}}{\|e_{I_i}\|\cdot\|e_{I_j}\|},\vspace{-1mm}
\end{equation}
where $e_{I_i}$ and $e_{I_j}$ denote embeddings of images $I_i$ and $I_j$. We then construct the distribution of the computed similarity scores $\mathcal{X}_s \sim sim_{cos}(e_i, e_j)$, $\forall (I_i, I_j) \in G$, by sorting all the pairs according to their calculated similarity score. From the distribution $\mathcal{X}_s$ we compute for each image $I_i$ its average pair index,
\vspace{-1mm}
\begin{equation}
    \overline{id^s_{I_i}} = \frac{1}{|\mathcal{I}|} \sum_{\mathcal{I}} id^{\mathcal{X}_s}(I_i, I_j),
\end{equation}
where $id^{\mathcal{X}_s}(\cdot)$ is a function, that for a given pair $(I_i, I_j)$ returns the index of $sim_{cos}(e_i, e_j)$ within the similarity distribution $\mathcal{X}_s$ and $\mathcal{I}$ represents the set of all image pairs $(I_i, I_j)$, where the quality $q_{I_i}$ is lower then $q_{I_j}$. The latter follows from the fact that the quality of an image pair is computed as $q(I_i, I_j) = min(I_i, I_j)$, i.e., it depends only on the image with the lower quality.
In addition, we construct a quality score distribution $\mathcal{X}_q \sim \{q_{I_i}\}^N_{i=1}$, by sorting the quality scores of all images within the given database. The average pair indices and the distribution $\mathcal{X}_q$ are then used to compute the optimized quality indices  
\vspace{-1mm}
\begin{equation}
    id^{\mathcal{X}_q}(q^*_{I_i}) = id^{\mathcal{X}_q}(q_{I_i}) + \theta \cdot (\overline{id^s_{I_i}} - id^{\mathcal{X}_q}(q_{I_i})),
\end{equation}
where $\theta$ is an open hyperparameter that controls the degree of change for the indices, and $id^{\mathcal{X}_q}(\cdot)$ is a function that returns, for some quality $q$, its index within the distribution $\mathcal{X}_q$. 

\textbf{Final steps.}
To avoid bias from randomly selecting mated pairs, we also repeat the entire process $R$ times, and average the final optimized quality indices, $\overline{id(q^*_{I_i})} = \frac{1}{R} \sum_{r=1}^R id^{\mathcal{X}_q}_r(q^*_{I_i})$ for all images. The images are then sorted by the calculated optimized quality indices $\overline{id(q^*_{I_i})}$ and assigned the quality score according to the output of the sorted list and the original quality score distribution $\mathcal{X}_q$.

\subsection{Transfer Learning}

One of the main goals of FIQA techniques is to improve the stability and performance of FR systems. We propose to use a pre-trained state-of-the-art FR model for quality prediction, as it efficiently extracts identity information from given facial images. Moreover, the embeddings generated by state-of-the-art FR models already contain some information about the quality of the input image. Formally, from an FR model $M$, we construct a quality regression model $H \circ M$, where $H$ represents a regression head. The regression head $H$ attempts to extract the quality of the input image $q_i = H(e_{I_i})$ from the embedding $e_{I_i}=M({I_i})$ and is learned through an $L_1$ loss applied over the optimized labels. To improve the transfer-learning process, we normalize the optimized quality scores to the interval of $[0, 1]$. 

\begin{figure}[!t]
    \centering
    \vspace{-1mm}\includegraphics[width=0.9\linewidth]{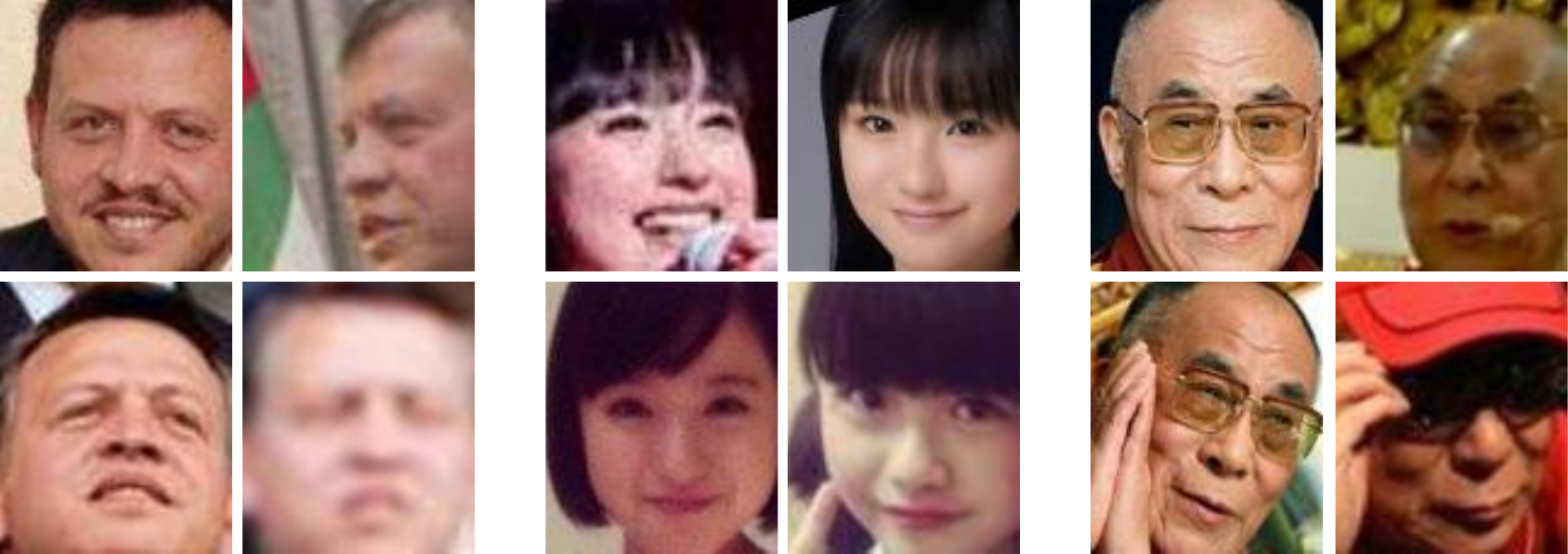}\vspace{-1mm}
    \caption{\textbf{Example VGGFace2 images.} Images of three distinct individuals are shown, illustrating the amount of variability present in the database. 
    \vspace{-2mm}}
    \label{fig:img_examples}
\end{figure}

\section{Experiments and Results}

\subsection{Experimental Setup}

\textbf{Training Database.} To train the proposed approach, a large-scale database of diverse facial images with rich appearance variability is needed. 
To this end, we select the VGGFace2 database~\cite{vggface2}, which contains over $3$ million images of more than $9000$ individuals. Images in the database vary in terms of facial pose, lighting conditions, image resolution, occlusions, and other similar factors that greatly affect the overall quality and appearance of the facial images, as also illustrated in Fig. \ref{fig:img_examples} for three individuals (in columns) from the database.

\textbf{Evaluation Setting.} We use six state-of-the-art FIQA methods as baselines to evaluate the proposed optimization scheme, i.e., CR-FIQA \cite{crfiqa}, FaceQAN \cite{faceqan}, MagFace \cite{magface}, SER-FIQ \cite{serfiq}, SDD-FIQA \cite{sddfiqa} and PCNet \cite{pcnet}. The baselines and the learned quality-regression networks are evaluated on five commonly used benchmark databases: XQLFW~\cite{xqlfw}, CPLFW~\cite{cplfw}, CFP-FP~\cite{cfp-fp}, CALFW~\cite{calfw} and LFW~\cite{lfw}. 
As the pre-trained FR model, we use ArcFace~\cite{arcface} with a ResNet100 architecture, trained on the MS1MV3 database using an angular margin loss. For the performance evaluation we consider two different scenarios: $(i)$ the {\textit{same-model scenario}}, where we use the ArcFace model for both quality-score prediction and generation of the performance indicators, and  $(ii)$ the {\textit{cross-model scenario}} where ArcFace is used for quality assessment, and the CurricularFace~\cite{curricularface} and ElasticFace-Cos+~\cite{elasticface} models are utilized to evaluate performance. Both of the test models are based on the ResNet100 architecture, but CurricularFace was trained on MS1MV2, while ElasticFace was trained with CASIA-WebFaces and MS1MV2.

\textbf{Performance Evaluation.} 
The performance of a FIQA technique directly correlates with its ability to properly rank images of similar quality. 
Therefore, to evaluate our approach, we follow standard FIQA evaluation methodology and use Error-versus-Reject-Characteristic (ERC) curves as the basis for performance reporting~\cite{surveypaper, faceqan, magface, crfiqa}. 
ERC curves measure the False Non-Match Ra\-te (FNMR) at a predefined False Match Rate (FMR), typically fixed at $0.001$, at various low-quality image drop (also unconsidered) rates. Specifically, we report the Area Under the ERC Curves (AUC) as our main performance indicator, where smaller values indicate better performance.

\textbf{Implementation Details.} When clustering the embedding space of each individual within the VGGFace2 database, we decide to set the number of clusters $C$ to $20$. Consequently, we generate $C - 1 = 19$, mated image pairs for each image, which means that each individual list of mated pairs consists of approximately $60$ million pairs. For the hyperparameter $\theta$ we use a relatively small value of $0.001$, since the goal is to optimize the already computed baseline quality scores. We repeat the whole process $10$ times and average the final results.
\begin{table}[t]
    \centering
    \caption{\textbf{Same-model performance comparison.} Comparison of AUC scores between baseline FIQA methods~(\textit{Baseline)} and our proposed optimization approach~(\textit{Optimized}) using ArcFace for quality estimation and performance evaluation. 
    Best results are highlighted in \begin{tabular}{c}\textbf{green.}\cellcolor{green!10}\end{tabular}}\vspace{-2mm}
    \label{tab:same_model_auc}
    \renewcommand{\arraystretch}{1.1}
    \resizebox{\columnwidth}{!}{%
    \begin{tabular}{ c c | c c c c c }
    
    \toprule
    \multicolumn{2}{c|}{\multirow{2}{*}{\textbf{Methods}}} & \multicolumn{5}{c}{\textbf{AUC@FMR1e-3$[\times 10^{-3}](\downarrow)$}} \\ \cline{3-7}

     &  &  LFW & CFP-FP & CPLFW & CALFW & XQLFW \\
    \midrule
    
    \multirow{2}{*}{CR-FIQA} &  Baseline &  \cellcolor{green!10}$\mathbf{1.7}$ & $1.3$ & \cellcolor{green!10}$\mathbf{86.5}$ & \cellcolor{green!10}$\mathbf{73.3}$ & $115.7$ \\
                            \cline{2-2}
                             &  Optimized & $1.8$ &  \cellcolor{green!10}$\mathbf{1.3}$ & $109.3$ & $73.7$ & \cellcolor{green!10}$\mathbf{105.7}$ \\
                                                                        
    \hline
    \multirow{2}{*}{FaceQAN} &  Baseline & $1.5$ & $1.9$ & $112.6$ & \cellcolor{green!10}$\mathbf{72.1}$ & $134.6$ \\
                            \cline{2-2}
                             &  Optimized & \cellcolor{green!10}$\mathbf{1.2}$ & \cellcolor{green!10}$\mathbf{1.8}$ & \cellcolor{green!10}$\mathbf{82.8}$ & $73.9$ & \cellcolor{green!10}$\mathbf{98.5}$\\
                                                                        
    \hline
    \multirow{2}{*}{SER-FIQ} &  Baseline & $2.9$ & $2.7$ & \cellcolor{green!10}$\mathbf{100.0}$ & $73.0$ & \cellcolor{gray!10}          
                                \\
                            \cline{2-2}
                             &  Optimized & \cellcolor{green!10}$\mathbf{2.4}$ & \cellcolor{green!10}$\mathbf{2.0}$ & $102.2$ & \cellcolor{green!10}$\mathbf{66.6}$ & \multirow{-2}{*}{\cellcolor{gray!10}$/$}
                             \\

    \hline
    \multirow{2}{*}{PCNet} &  Baseline & $1.7$ & $2.3$ & $88.5$ & \cellcolor{green!10}$\mathbf{59.3}$ & $139.7$ \\
                            \cline{2-2}
                             &  Optimized & \cellcolor{green!10}$\mathbf{0.9}$ & \cellcolor{green!10}$\mathbf{1.7}$ & \cellcolor{green!10}$\mathbf{87.6}$ & $60.2$ & \cellcolor{green!10}$\mathbf{116.4}$ \\
    
    \hline
    \multirow{2}{*}{MagFace} &  Baseline & $2.0$ & $3.7$ & \cellcolor{green!10}$\mathbf{118.6}$ & $74.4$ & $294.9$ \\
                            \cline{2-2}
                             &  Optimized & \cellcolor{green!10}$\mathbf{1.7}$ & \cellcolor{green!10}$\mathbf{1.7}$ & $125.3$ & \cellcolor{green!10}$\mathbf{72.3}$ & \cellcolor{green!10}$\mathbf{127.6}$\\
    
    \hline
    \multirow{2}{*}{SDD-FIQA} &  Baseline & $1.8$ & $2.6$ & $91.4$ & $87.5$ & $190.1$ \\
                            \cline{2-2}
                             &  Optimized & \cellcolor{green!10}$\mathbf{1.8}$ & \cellcolor{green!10}$\mathbf{1.7}$ & \cellcolor{green!10}$\mathbf{80.6}$ & \cellcolor{green!10}$\mathbf{66.6}$ & \cellcolor{green!10}$\mathbf{135.3}$ \\
                             
    \bottomrule
    
    \end{tabular}
    }
    \vspace{-3mm}
\end{table}

\begin{table*}[t]
    \centering
    \caption{\textbf{Cross-model performance comparison.} Comparison of AUC scores (AUC@FMR1e-3$[\times 10^{-3}](\downarrow)$) 
    between the baseline FIQA methods~(\textit{Baseline)} and the proposed optimization approach~(\textit{Optimized}), using the ArcFace model  for quality estimation and ElasticFace and CurricularFace for performance scoring. 
    Best results are highlighted in \begin{tabular}{c}\textbf{green}\cellcolor{green!10}\end{tabular}.}\vspace{-2mm}
    \label{tab:cross_model_auc}
    \renewcommand{\arraystretch}{1.07}
    \resizebox{0.8\textwidth}{!}{%
    \begin{tabular}{ c c | c c c c c | c c c c c  }
    
    \toprule
    \multicolumn{2}{c|}{\multirow{2}{*}{\textbf{Methods}}}& \multicolumn{5}{c|}{\textbf{ElasticFace}} & \multicolumn{5}{c}{\textbf{CurricularFace}} \\ \cline{3-12}
     & & LFW & CFP-FP & CPLFW & CALFW & XQLFW & LFW & CFP-FP & CPLFW & CALFW & XQLFW \\
    \midrule
    
    \multirow{2}{*}{CR-FIQA} &  Baseline & \cellcolor{green!10}$\mathbf{2.1}$ & $2.1$ & \cellcolor{green!10}$\mathbf{62.0}$ & \cellcolor{green!10}$\mathbf{74.9}$ & $315.8$ & \cellcolor{green!10}$\mathbf{1.8}$ & $1.9$ & $57.3$ & \cellcolor{green!10}$\mathbf{76.6}$ & $281.0$\\
                            \cline{2-2}
                             &  Optimized & $2.3$ & \cellcolor{green!10}$\mathbf{2.0}$ & $62.7$ & $76.1$ & \cellcolor{green!10}$\mathbf{305.5}$ & $2.0$ & \cellcolor{green!10}$\mathbf{1.9}$ & \cellcolor{green!10}$\mathbf{56.1}$ & $78.2$ & \cellcolor{green!10}$\mathbf{272.2}$ \\
                                                                        
    \hline
    \multirow{2}{*}{FaceQAN} &  Baseline & \cellcolor{green!10}$\mathbf{1.9}$ & \cellcolor{green!10}$\mathbf{2.6}$ & $66.4$ & \cellcolor{green!10}$\mathbf{77.7}$ & $329.6$ & $1.6$ & \cellcolor{green!10}$\mathbf{2.2}$ & $70.7$ & \cellcolor{green!10}$\mathbf{82.6}$ & $323.3$ \\
                            \cline{2-2}
                             &  Optimized & $2.0$ & $2.7$ & \cellcolor{green!10}$\mathbf{65.0}$ & $79.7$ & \cellcolor{green!10}$\mathbf{287.1}$ & \cellcolor{green!10}$\mathbf{1.5}$ & $2.3$ & \cellcolor{green!10}$\mathbf{63.7}$ & $83.7$ & \cellcolor{green!10}$\mathbf{263.5}$ \\
                                                                        
    \hline
    \multirow{2}{*}{SER-FIQ} &  Baseline & $3.8$ & $3.9$ & $79.4$ & $74.9$ & \cellcolor{gray!10} 
                                         & $3.2$ & $3.4$ & $69.6$ & $78.1$ & \cellcolor{gray!10} 
                                         \\
                            \cline{2-2}
                             &  Optimized & \cellcolor{green!10}$\mathbf{3.2}$ & \cellcolor{green!10}$\mathbf{3.0}$ & \cellcolor{green!10}$\mathbf{77.5}$ & \cellcolor{green!10}$\mathbf{69.7}$ & \multirow{-2}{*}{\cellcolor{gray!10}$/$} 
                             & \cellcolor{green!10}$\mathbf{2.5}$ & \cellcolor{green!10}$\mathbf{2.5}$ & \cellcolor{green!10}$\mathbf{69.2}$ & \cellcolor{green!10}$\mathbf{71.8}$ & \multirow{-2}{*}{\cellcolor{gray!10}$/$} 
                             \\
    
    \hline
    \multirow{2}{*}{PCNet} &  Baseline & $3.2$ & $3.5$ & $74.4$ & $62.0$ & $312.2$ & $2.1$ & $3.0$ & $66.5$ & \cellcolor{green!10}$\mathbf{64.3}$ & $300.3$ \\
                            \cline{2-2}
                             &  Optimized & \cellcolor{green!10}$\mathbf{2.0}$ & \cellcolor{green!10}$\mathbf{2.5}$ & \cellcolor{green!10}$\mathbf{69.5}$ & \cellcolor{green!10}$\mathbf{62.7}$ & \cellcolor{green!10}$\mathbf{295.2}$ & \cellcolor{green!10}$\mathbf{1.3}$ & \cellcolor{green!10}$\mathbf{2.3}$ & \cellcolor{green!10}$\mathbf{62.0}$ & $64.7$ & \cellcolor{green!10}$\mathbf{272.6}$ \\
    
    \hline
    \multirow{2}{*}{MagFace} &  Baseline & $2.6$ & $4.9$ & $79.5$ & $74.9$ & $601.8$ & $2.1$ & $4.4$ & $81.5$ & $79.9$ & $593.2$ \\
                            \cline{2-2}
                             &  Optimized & \cellcolor{green!10}$\mathbf{2.1}$ & \cellcolor{green!10}$\mathbf{2.5}$ & \cellcolor{green!10}$\mathbf{68.6}$ & \cellcolor{green!10}$\mathbf{73.9}$ & \cellcolor{green!10}$\mathbf{306.6}$ & \cellcolor{green!10}$\mathbf{2.0}$ & \cellcolor{green!10}$\mathbf{2.2}$ & \cellcolor{green!10}$\mathbf{65.1}$ & \cellcolor{green!10}$\mathbf{77.6}$ & \cellcolor{green!10}$\mathbf{260.4}$\\
    
    \hline
    \multirow{2}{*}{SDD-FIQA} &  Baseline & $3.1$ & $3.8$ & $73.1$ & $79.9$ & $480.8$ & $2.1$ & $3.3$ & $64.8$ & $77.4$ & $438.6$ \\
                            \cline{2-2}
                             &  Optimized & \cellcolor{green!10}$\mathbf{2.5}$ & \cellcolor{green!10}$\mathbf{2.5}$ & \cellcolor{green!10}$\mathbf{63.4}$ & \cellcolor{green!10}$\mathbf{68.7}$ & \cellcolor{green!10}$\mathbf{292.8}$ & \cellcolor{green!10}$\mathbf{2.0}$ & \cellcolor{green!10}$\mathbf{2.1}$ & \cellcolor{green!10}$\mathbf{55.7}$ & \cellcolor{green!10}$\mathbf{70.4}$ & \cellcolor{green!10}$\mathbf{268.3}$\\
                             
    \bottomrule
    
    \end{tabular}
    }
    \vspace{-1mm}
\end{table*}

\subsection{Results}

Before presenting results, we note that SER-FIQ was used in the construction of the XQLFW database, so any results that combine the two are excluded from the presented analysis. 

\begin{figure*}[!t]
    \centering
    \includegraphics[width=0.9\textwidth]{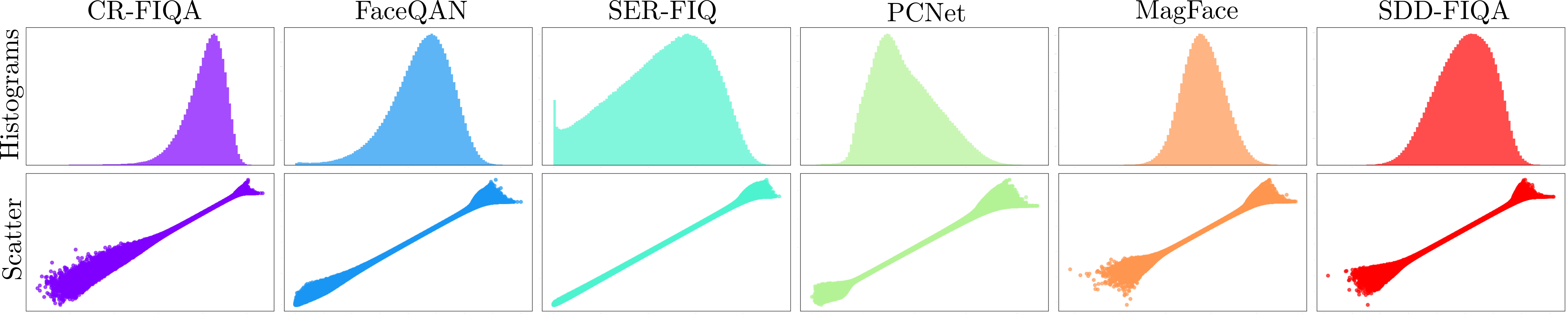}\vspace{-2mm}
    \caption{\textbf{Qualitative analysis of the proposed approach.} For each FIQA method, we show the prior distribution of the quality scores of the VGGFace2 database, and an associated scatter plot showing the changes in the quality scores due to our optimization approach.}\vspace{-1mm}
    \label{fig:qualitative_analysis}
\end{figure*}

\textbf{Same-Model Results.} Table~\ref{tab:same_model_auc} shows the AUC values produced directly with the original FIQA methods (labeled \textit{Baseline}) as well as the AUC scores of the quality-regression network trained using our optimized labels (marked \textit{Optimized}). For readability purposes, the AUC scores are multiplied by $10^3$ and rounded to one decimal place. We observe that in most cases the results of our approach are better than those of the underlying FIQA approaches. The only exception to this observation is CR-FIQA, where a concrete improvement is observed only for the hardest of the considered datasets, i.e., XQLFW, while the results for the remaining datasets are mostly close, but deteriorate drastically for CPLFW. For all other methods the results consistently improve, with occasional outliers on the CALFW or CPLFW benchmarks.

\textbf{Cross-Model Results.}
Table~\ref{tab:cross_model_auc} again shows the AUC values of both the baseline and our (optimized) regression-based FIQA techniques, but this time computed for the cross-model scenario, where the FR model used for estimating the quality of the input images differs from the FR model used for performance reporting. 
Looking at the individual methods, CR-FIQA and FaceQAN do not show a clear edge for either the baseline or optimized results. While for the hardest benchmark, XQLFW, the optimized variant always performs better than the baseline variant, the opposite is true for CALFW, which contains cross-age image data. For all other FIQA approaches, the proposed optimization method yields better results, and outperforms the baselines in all cases except for PCNet on CALFW. The results are consistent for both the ElasticFace and CurricularFace model. 

\textbf{Cross-Model vs. Same-Model Results.}
Comparing the cross-model with the same-model results, many similarities can be observed. The performance benefit due to the optimization approach is relatively unconvincing for CR-FIQA, while the results for all other methods are mirrored between the two evaluation schemes. The biggest difference is seen for FaceQAN, where the proposed method performs comparably worse in the cross-model evaluation setting. 

\textbf{Qualitative Analysis.} If we look more closely at how the proposed approach works, we see that the distribution of the initial quality scores remains the same under the optimization scheme. This is because the method only rearranges the order of the images and assigns them quality scores from the prior distribution. 
However, a potential problem with this approach, is that the quality scores of images in higher density areas of the distribution, are harder to change than the quality scores of the images in the lower density areas.  This phenomenon is well illustrated in Fig.~\ref{fig:qualitative_analysis}, where for each of the FIQA methods used, a histogram of the prior quality scores over VGGFace2  is presented together with a scatter plot, where each point represents the prior quality of a given image on the $x$-axis and the optimized quality score on the $y$-axis. 
Note how the quality scores in areas of lower density seem to change drastically, while almost no movement is observed in higher density areas.

\begin{table*}[!t]
    \centering
    \caption{\textbf{Results of the ablation study.} Shown are AUC scores (AUC@FMR1e-3$[\times 10^{-3}](\downarrow)$) generated by the regression-based FIQA methods trained with the initial/prior quality labels and the performance gain(-)/loss(+) in terms of AUC change due to the label optimization procedure (in brackets). Results are presented for the same-model (ArcFace) and cross-model (ElasticFace and CurricularFace) settings. Gains are marked green, losses red.}\vspace{-2mm}
    \renewcommand{\arraystretch}{1.12}
    \label{tab:ablation_study}
    \resizebox{0.9\textwidth}{!}{%
    \begin{tabular}{ l | l l l | l l l | l l l  }
    
    \toprule
    \multicolumn{1}{c|}{\multirow{2}{*}{\textbf{Methods}}} & \multicolumn{3}{c|}{\textbf{ArcFace}} & \multicolumn{3}{c|}{\textbf{ElasticFace}} & \multicolumn{3}{c}{\textbf{CurricularFace}} \\ \cline{2-10}
     & \multicolumn{1}{c}{LFW} & \multicolumn{1}{c}{CPLFW} & \multicolumn{1}{c |}{XQLFW} & \multicolumn{1}{c}{LFW} & \multicolumn{1}{c}{CPLFW} & \multicolumn{1}{c |}{XQLFW}  & \multicolumn{1}{c}{LFW} & \multicolumn{1}{c}{CPLFW} & \multicolumn{1}{c}{XQLFW}   \\
    \midrule
    
    \multirow{1}{*}{CR-FIQA} & \cellcolor{red!10}$1.3~(+0.5)$ & \cellcolor{red!10}$84.0~(+1.7)$ & \cellcolor{red!10}$102.8~(+2.9)$ & \cellcolor{red!10}$2.1~(+0.2)$ & \cellcolor{green!10}$65.0~(-2.3)$ & \cellcolor{red!10}$280.4~(+24.9)$ & \cellcolor{red!10}$1.6~(+0.4)$ & \cellcolor{green!10}$64.2~(-4.1)$ & \cellcolor{green!10}$273.3~(-1.1)$\\

    \multirow{1}{*}{FaceQAN} & \cellcolor{green!10}$1.4~(-0.2)$ & \cellcolor{green!10}$85.7~(-2.9)$ & \cellcolor{green!10}$109.8~(-11.3)$ & \cellcolor{green!10}$2.1~(-0.1)$ & \cellcolor{green!10}$65.5~(-0.5)$ & \cellcolor{green!10}$298.2~(-11.1)$ & \cellcolor{green!10}$1.7~(-0.2)$ & \cellcolor{green!10}$65.2~(-1.5)$ & \cellcolor{green!10}$273.0~(-9.5)$\\

    \multirow{1}{*}{SER-FIQ} & \cellcolor{red!10}$2.2~(+0.2)$ & \cellcolor{red!10}$101.9~(+0.3)$ & \multicolumn{1}{c}{\cellcolor{gray!10}$/$}
    & \cellcolor{green!10}$3.2~(-0.0)$ & \cellcolor{green!10}$79.5~(-2.0)$ & \multicolumn{1}{c}{\cellcolor{gray!10}$/$}
    & \cellcolor{red!10}$2.4~(+0.1)$ & \cellcolor{green!10}$72.8~(-3.6)$ & \multicolumn{1}{c}{\cellcolor{gray!10}$/$}
    \\

    \multirow{1}{*}{PCNet} & \cellcolor{green!10}$0.9~(-0.0)$ & \cellcolor{green!10}$93.8~(-6.2)$ & \cellcolor{green!10}$211.3~(-94.9)$ & \cellcolor{green!10}$2.5~(-0.5)$ & \cellcolor{green!10}$78.6~(-9.1)$ & \cellcolor{green!10}$406.6~(-111.4)$ & \cellcolor{red!10}$1.2~(+0.1)$ & \cellcolor{green!10}$69.2~(-7.2)$ & \cellcolor{green!10}$360.1~(-87.5)$\\
    
    \multirow{1}{*}{MagFace} & \cellcolor{green!10}$1.8~(-0.1)$ & \cellcolor{red!10}$117.2~(+8.1)$ & \cellcolor{green!10}$195.5~(-67.9)$ & \cellcolor{red!10}$2.0~(+0.1)$ & \cellcolor{green!10}$69.8~(-1.2)$ & \cellcolor{green!10}$492.3~(-185.7)$ & \cellcolor{green!10}$2.1~(-0.1)$ & \cellcolor{green!10}$66.1~(-1.0)$ & \cellcolor{green!10}$501.8~(-241.4)$\\
    
    \multirow{1}{*}{SDD-FIQA} & \cellcolor{green!10}$2.2~(-0.4)$ & \cellcolor{green!10}$80.9~(-0.3)$ & \cellcolor{green!10}$164.8~(-29.5)$ & \cellcolor{green!10}$2.9~(-0.4)$ & \cellcolor{green!10}$64.3~(-0.9)$ & \cellcolor{green!10}$393.8~(-101.0)$ & \cellcolor{green!10}$2.4~(-0.4)$ & \cellcolor{green!10}$57.5~(-1.8)$ & \cellcolor{green!10}$337.1~(-68.8)$\\
                             
    \bottomrule
    
    \end{tabular}
    }
    \vspace{-3mm}
\end{table*}

\textbf{Ablation Study.}
To demonstrate how the optimization of the quality labels affects the final results, we present in 
Table~\ref{tab:ablation_study} AUC scores obtained with a quality-regression network trained with the initial (unoptimized) quality labels as well as the performance gain({\color{green!90}-})/loss({\color{red!90}+}) 
due to the optimization procedure (in brackets). We use the two most difficult benchmarks: CPLFW and XQLFW, as well as the LFW benchmark for this ablation study. From the presented  results, we see that the effectiveness of the optimization in the \textit{same-model scenario}, i.e. with ArcFace, to a certain extent depends on the chosen FIQA technique. For CR-FIQA and SER-FIQ the results do not really seem to favour the optimization approach, as most of the performance gains observed in Table~\ref{tab:same_model_auc} appear to be a consequence of the transfer learning step. On the \textit{cross-model} side, the results for both ElasticFace and CurricularFace seem to be more in favour of the optimized labels, with only a few counterexamples on the LFW database.

\textbf{Run-time performance.} Because we use a regression-based model trained with the optimized quality scores, the run-time performance of our approach is (approximately) the same regardless of the initial FIQA method used as the basis for the reference quality scores. Thus, the proposed transfer learning step can also be seen as a knowledge distillation procedure that allows us to retain the performance of a given FIQA technique while ensuring a (approximately) fixed run-time complexity, as evidenced by the run-times in Table~\ref{tab:time_complexity} - computed on a desktop PC with an Intel i9-10900KF ($3.70$GHz) CPU and a Nvidia 3090 GPU with $24$GB of video RAM. 

\begin{table}[!t]
    \centering
    \caption{\textbf{Run-time performance.} Shown in [$\mu$s].}\vspace{-2mm}
    \label{tab:time_complexity}
    \resizebox{\columnwidth}{!}{%
    \begin{tabular}{ l | l l l l l l  } 
    \toprule
   Method & CR-FIQA & FaceQAN & SER-FIQ & PCNet & MagFace & SDD-FIQA \\
   \midrule
   Original & $11.5\pm5.0$ & $346.5\pm9.0$ & $125.2\pm19.5$ & $14.5\pm1.7$ & $11.5\pm4.9$ & $5.7\pm5.0$ \\ 
   Ours & $11.4\pm3.0$ & $11.3\pm5.0$ & $11.3\pm5.0$ & $10.6\pm4.9$ & $11.3\pm5.0$ & $11.3\pm5.0$ \\
   \midrule
   Speed-up$^\dagger$ & \cellcolor{green!10}$+0.8\%$ & \cellcolor{green!10}$+3000\%$ & \cellcolor{green!10}$+1100\%$ & \cellcolor{green!10}$+36\%$ & \cellcolor{green!10}$+1.7\%$ & \cellcolor{red!10}$-50\%$ \\                       
    \bottomrule
     \multicolumn{7}{l}{\footnotesize $^\dagger$ Approximate values.}\\
    \end{tabular}
    }
    \vspace{-5mm}
\end{table}

\section{Conclusion}


We presented a novel optimization approach, that aims to improve the performance of modern FIQA approaches. 
A thorough evaluation was performed using multiple state-of-the-art FIQA methods, datasets and FR models. 
The results of the evaluation showed significant performance improvements in most cases when using the optimization scheme both in the same-model and cross-model setting. 
As part of our future work, we plan to incorporate multiple sources of quality scores into the optimization procedure to benefit from the complementary quality description provided by different FIQA techniques.

\bibliographystyle{ieeetr}
\bibliography{bibliography}

\end{document}